\pdfoutput=1
\documentclass[11pt]{article}
\usepackage{authblk}
\usepackage[preprint]{acl}
\usepackage{times}
\usepackage{latexsym}
\usepackage[T1]{fontenc}
\usepackage[utf8]{inputenc}
\usepackage{microtype}
\usepackage{inconsolata}
\usepackage{enumerate}
\usepackage{enumitem}
\usepackage{graphicx}
\usepackage{tabularx}
\usepackage{booktabs}
\usepackage{etoolbox}
\newtoggle{appendixes}
\toggletrue{appendixes}

\usepackage{tikz}
\usepackage{xspace}
\usepackage[normalem]{ulem}
\usepackage{tabulary}
\usepackage{tabularx}
\usepackage{array}

\definecolor{red-bf}{RGB}{255,99,132}  
\definecolor{blue-bf}{RGB}{54,162,235}  
\definecolor{yellow-bf}{RGB}{255,206,86}  
\definecolor{greenblue-bf}{RGB}{75,192,192}  
\definecolor{purple-bf}{RGB}{153,102,255} 
\definecolor{darkred-bf}{RGB}{205,89,112}  
\definecolor{darkblue-bf}{RGB}{44,142,215}  
\definecolor{darkpurple-bf}{RGB}{133,92,225}  

\newcommand{\ours}{\textsc{FELT}\xspace}

\definecolor{learner_blue}{rgb}{0.28235, 0.57647, 0.7098}
\definecolor{task_green}{rgb}{0.30588, 0.6549, 0.18039}
\definecolor{error_purple}{rgb}{0.62745, 0.16863, 0.57647}
\definecolor{feedback_orange}{rgb}{0.91373, 0.44314, 0.19608}

\usepackage{soul}
\newcommand{\eg}{e.g.,\xspace}
\newcommand{\ie}{i.e.,\xspace}

\newcommand{\squishlist}{
  \begin{list}{$\bullet$}
    { \setlength{\itemsep}{0pt}      \setlength{\parsep}{3pt}
      \setlength{\topsep}{3pt}       \setlength{\partopsep}{0pt}
      \setlength{\leftmargin}{1.5em} \setlength{\labelwidth}{1em}
      \setlength{\labelsep}{0.5em} } }
\newcommand{\reallysquishlist}{
  \begin{list}{$\bullet$}
    { \setlength{\itemsep}{0pt}    \setlength{\parsep}{0pt}
      \setlength{\topsep}{0pt}     \setlength{\partopsep}{0pt}
      \setlength{\leftmargin}{0.2em} \setlength{\labelwidth}{0.2em}
      \setlength{\labelsep}{0.2em} } }

 \newcommand{\squishend}{
     \end{list} 
 }

\renewcommand{\cite}{\citep}

\title{Let Me Teach You: \\ Pedagogical Foundations of Feedback for Language Models}

\author[1]{\textbf{Beatriz Borges}}
\author[2]{\textbf{Niket Tandon}}
\author[1]{\textbf{Tanja Käser}}
\author[1]{\textbf{Antoine Bosselut}}
\affil{EPFL \:\:\: $^2$Allen Institute for Artificial Intelligence}
\affil[ ]{\texttt{\{beatriz.borges, antoine.bosselut\}@epfl.ch}}

\begin{document}
\maketitle
\begin{abstract}
Natural Language Feedback (NLF) is an increasingly popular mechanism for aligning Large Language Models (LLMs) to human preferences. Despite the diversity of the information it can convey, NLF methods are often hand-designed and arbitrary, with little systematic grounding. At the same time, research in learning sciences has long established several effective feedback models. 
In this opinion piece, we compile ideas from pedagogy to introduce \ours, a feedback framework for LLMs that outlines various characteristics of the feedback space, and a feedback content taxonomy based on these variables, providing a general mapping of the feedback space.
In addition to streamlining NLF designs, \ours also brings out new, unexplored directions for research in NLF. We make our taxonomy available to the community, providing guides and examples for mapping our categorizations to future research.
\end{abstract}

\section{Introduction}

The last few years introduced a new paradigm for finetuning Large Language Models (LLMs): learning from human feedback \citep{ziegler2020finetuning, stiennon_learning_2022, bai2022training, openai2023gpt4, rafailov2023direct, azar2023general, fisch2024robust} to augment their capabilities beyond their pretraining \citep{christiano2017deeprl, wu2021recursively, menick2022teaching}. This alignment has yielded less toxic and harmful models \citep{bai_constitutional_2022, korbak2023pretraining} that are preferred by users \citep{ouyang2022instructGPT, bai_constitutional_2022}.

\begin{figure}[t]
\centering
\includegraphics[width=1.0\columnwidth]{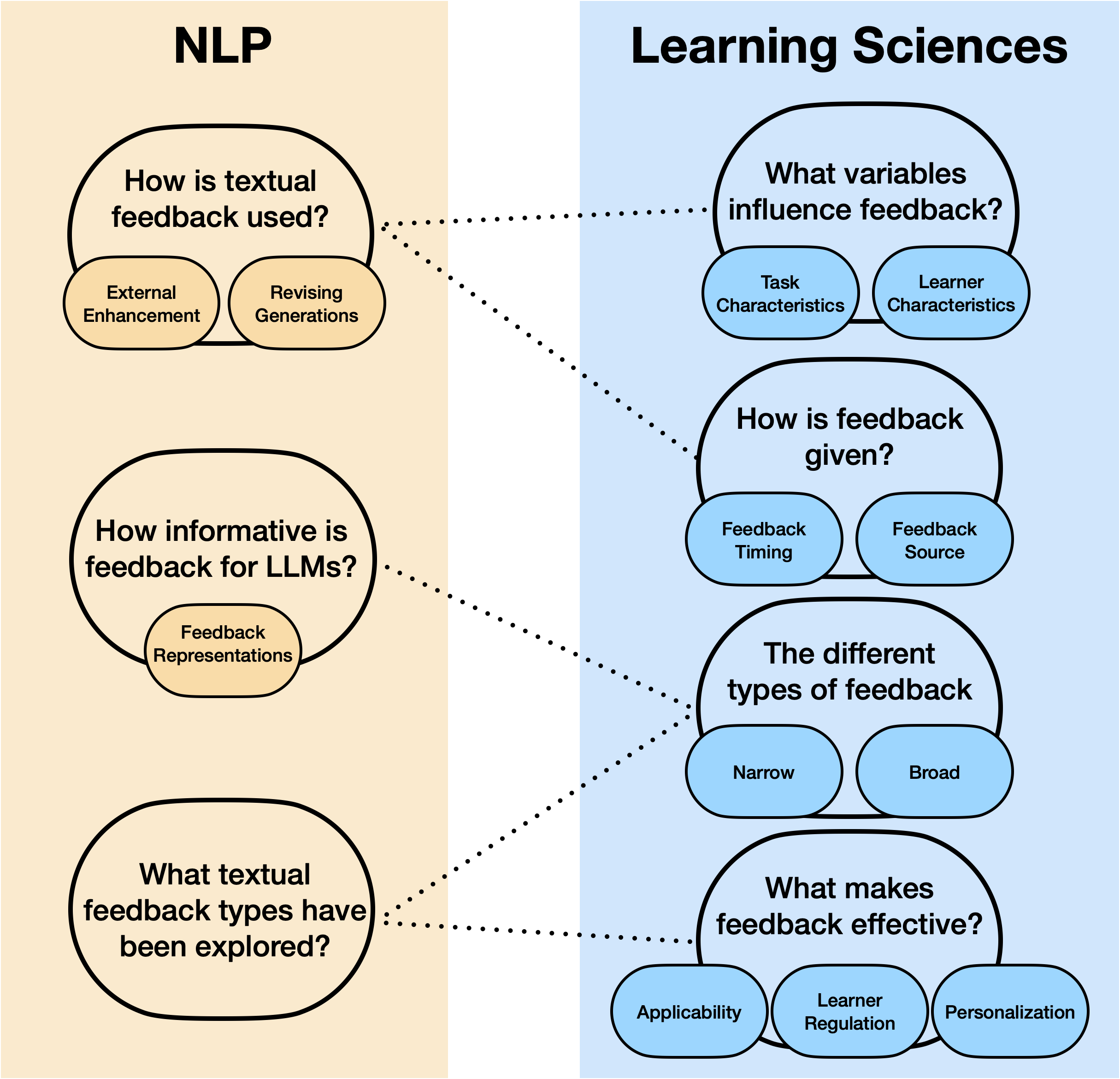}
\caption{Connecting feedback research in NLP to foundations of feedback in the Learning Sciences. 
}
\label{fig:pedagogy_nlp_relation}
\end{figure}

Whether learning from feedback is done by directly learning from human preferences \citep{rafailov2023direct, azar2023general, hong2024orpo, meng2024simpo, saeidi2024triple, fisch2024robust}, or with Reinforcement Learning from Feedback (RLF) using human-curated (RLHF; \citealp{ouyang2022instructGPT, bai2022training, ramamurthy2023reinforcement}) or AI-generated (RLAIF; \citealp{bai_constitutional_2022, saunders2022selfcritiquing, madaan2023selfrefine}) feedback, all variants have been shown to be successful in several metrics --- from encouraging honest behaviors, to reducing toxicity, to being preferred by evaluators \citep{ouyang2022instructGPT}. Other approaches, such as imitation learning \citep{li2017dialogue, kreutzer-etal-2018-neural, hancock2019learning, scheurer2022training}, and feedback modeling \citep{weston_dialog-based_2016, li2017dialogue, hancock2019learning, xu2022learning, liu2023chain} have had similar success. Feedback has thus emerged as an important source of information for model improvement and evaluation, directing them toward desired objectives and behaviors \citep{fernandes_bridging_2023}.

Despite the observed benefits of learning from feedback, a systematic study of what constitutes helpful feedback has so far remained absent. For example, RLF requires a Reward Model (RM) to be trained on numerical or ranking-based feedback data \citep{rafailov2023direct} --- a format limited in the amount of information it conveys \citep{wu2023finegrained}. To counteract this limitation, works have begun exploring Natural Language Feedback (NLF; \citealp{weston_dialog-based_2016, madaan2023selfrefine, wu2023finegrained}). However, these works rely on ``intuitive guesses'' about what constitutes useful feedback --- leading different works to explore distinct conceptualizations of NLF, preventing a systematic comparison. 

To establish a more concrete foundation for learning from feedback, we survey comprehensive studies of feedback from the field of learning sciences, which investigate feedback as an essential component of instruction and learning. We unify feedback approaches for NLP into a common framework, grounded on our surveyed pedagogical foundations (as summarized in Figure~\ref{fig:pedagogy_nlp_relation}).
To this end, we first present the most relevant feedback-related models from the learning sciences (§\ref{sec:background_pedagogy}).
Taking inspiration from these pedagogical models, we create a novel framework, FELT, that expansively maps the various features of the LLM feedback space to pedagogical foundations (§\ref{sec:framework}), and identify both dimensions studied by previous works as well as several important aspects of feedback that remain underexplored. We then focus on the least explored one, \textit{feedback content}, and introduce a novel comprehensive taxonomy systematizing both the content and the delivery of natural language feedback (§\ref{sec:taxonomy}), shining a light on the underspecification of current approaches to feedback, and suggesting promising areas of future research.

\vspace{-2pt}
\paragraph{Contributions} We 
(i) present a survey of pedagogical feedback formulations and models;
(ii) organize the variables that influence feedback (and its processing) into a schematic framework, specifically adapted for LLMs; (iii) propose a general and extensive taxonomy of feedback content; and (iv) propose areas of future research based on gaps between our taxonomy and the current landscape of LLM research on feedback.

\section{Feedback in NLP}
\label{sec:background_nlp}
In this section, we survey current conceptualizations and applications of feedback in NLP, before exploring the feedback models and perspectives developed in the domain of learning science (§\ref{sec:background_pedagogy}).

\subsection{Current State of Feedback in NLP} 
The value of feedback is derived from the implicit knowledge it represents about human values and expectations, that would otherwise be extremely difficult to specify \citep{christiano2017deeprl}. 
Feedback can assume different forms: numerical ratings, rankings, preferences, demonstrations, and textual information (either using a template or free-form text --- structured and unstructured feedback, respectively). 
RLF methodologies usually collect either a numeric rating or a ranking for classifying the \textit{quality} of an initial answer (often encouraging properties such as \textit{helpfulness} and \textit{honesty} while mitigating \textit{harmfulness}; \citealp{askell2021general}). 
RLF may also leverage demonstrations to finetune LLMs in a supervised fashion before the RLF stage to reduce the search space \citep{ouyang2022instructGPT, bai_constitutional_2022}. RLF methods address the intractable problem of designing an appropriate loss function for training models to exhibit behaviors with no closed-form solution \citep{askell2021general}. Recent works have also started to incorporate feedback in a prompt-based manner \citep{schick2022peer, madaan2023selfrefine, paul2023refiner, chen2023teaching, lin2023unlocking, zhao2024incontext}.

\paragraph{Informativeness of Feedback} However, the extent to which feedback formulations transmit their goal states remains unclear.  
For example, InstructGPT \citep{ouyang2022instructGPT} is first finetuned on demonstration data, and then performs RLHF with a reward model trained using comparison data (\ie pairs of ranked generations). 
This feedback is limited in the information it transmits. Scoring a demonstration A as ``better'' than demonstration B provides little information on the quality of A nor B,
\footnote{We note such a format also obfuscates any bias and disagreement that occurred in reaching such a judgment.} nor on how either demonstration can be improved. Taking these limitations  into account, RMs are likely to suffer from some degree of misalignment \citep{pan2022effects, gao2022scaling, song2023reward}. 
Recent works have started to acknowledge the limited information in these feedback formulations, recognizing them as unsuited for capturing critical information, such as different types of errors \citep{golovneva2023roscoe, wu2023finegrained}.

\subsection{Natural Language Feedback for LLMs}

The most commonly used feedback formulations, scalar and ranking feedback, are thus limited in the information they convey, motivating new methods that leverage natural language for more expressive feedback formulations \citep{fernandes_bridging_2023}.

\paragraph{How should feedback be provided to LLMs?} Certain works augment the model through data augmentation \citep{shi2022life}, external corrective feedback \citep{tandon-etal-2022-learning, madaan-etal-2022-memory, shinn2023reflexion} or natural language patches \citep{murty2022fixing}. Another line of work introduces a secondary model, that either refines an original LLM's answer \citep{scheurer2022training, welleck2022generating, tandon-etal-2022-learning}, critiques it \citep{saunders2022selfcritiquing, paul2023refiner} iteratively self-improves \citep{schick2022peer, chen2023teaching, madaan2023selfrefine, yuan2024selfrewarding} or otherwise constrains the initial response of a model \citep{stephan2024rlvf}. 
Beyond increasing the complexity of feedback using natural language, several of these approaches also target intermediate generations with their feedback, not the final outcome produced by the model, thereby increasing the number of feedback opportunities through multiple iterations \citep{lightman2023lets}. 

\paragraph{What information should feedback content convey?}

\citet{shi2022life} distinguish textual feedback depending on whether the feedback is formally provided to the model's answer, or, remains in the dialog setting, where the user mentions they disliked the reply they received. 
\textsc{SELF-REFINE} \citep{madaan2023selfrefine} argues that the quality of the generated feedback is critical, though they only compare their ``actionable and specific'' LLM-generated feedback against ``generic feedback'' and no feedback at all.
\citet{wu2023finegrained} propose the introduction of finer-grained feedback --- and of three different error types: factual incorrectness, irrelevance, and information incompleteness. Despite its impressive performance, the feedback exploration in this work is limited at only three specific types, and only uses preference rankings.
Finally, \citet{weston_dialog-based_2016} conducted the most thorough exploration of NLF to date, with 10 different dialogue-based supervision modes, which represent different interaction and feedback types. However, these often overlap information-wise, limiting its conclusions.  

\vspace{-4pt}
\paragraph{What is missing?}
Various works conceptualize feedback in completely different ways. No work has taken up a true mapping of the feedback space, identified the different types of information that can be encoded in NLF, and allowed for an exploration of different feedback components and their effectiveness. To address this, we look to the learning sciences, which studies feedback as an integral component of human learning. We survey their different conceptualizations of feedback (\S\ref{sec:background_pedagogy}), and unify them under a new framework, \ours, adapted specifically for LLMs (\S\ref{sec:framework}).

\section{Feedback in Learning Sciences}
\label{sec:background_pedagogy}

To construct a comprehensive, theory-grounded model of feedback that addresses the limited exploration of NLF in NLP, we build off the work of \citet{lipnevich_review_2021}, who conducted a systematic review of 15 relevant and influential works on feedback models research in education. We provide a brief overview of the key points of each of these works and use them to subsequently propose a framework for the feedback ecosystem (\S\ref{sec:framework}) and a taxonomy for feedback content (\S\ref{sec:taxonomy}).

\subsection{What is feedback?}

Many works agree that feedback is, or contains, information provided to a learner. 
While studies may disagree on the breadth or specificity required for feedback, and other limitations, a definition (adopted throughout this paper) emerges from their consensus:\iftoggle{appendixes}{\footnote{For an overview of all the different definitions of feedback discussed, please see Appendix \ref{app:feedback_definitions}.} 
}{} \textit{feedback is any task-relevant information given to a learner (content), by any possible feedback-generating agent (source)}.

\subsection{What constitutes effective feedback?}
\label{sec:eff-feedback}

\citet{kluger_effects_1996} showed feedback was detrimental to a learner's performance in $38\%$ of analyzed cases. 
Three main requirements for helpful feedback emerge from previous work: \textit{applicability}, \textit{learner regulation}, and \textit{personalization}. These requirements are directly related to feedback content, which we explore in §\ref{sec:taxonomy} for LLMs.\footnote{{Looking forth, \textit{Applicability} corresponds to dimensions of \textit{applicability of instructions}, \textit{learner regulation} to \textit{purpose}, and \textit{personalization} is a product of the combination of all dimensions, allowing a fine-grained customization of feedback.}}

\vspace{-4pt}
\paragraph{Applicability} \textit{Feedback should be actionable, allowing the learner to achieve a desired target performance.} 
\citet{sadler_formative_1989} suggests that feedback needs to identify a target performance, compare the learner's current performance to it, and engage in actions to reduce that difference. Similarly, \citet{hattie_power_2007} indicate that effective feedback needs to answer three questions: where the learner is going (the goal), how they can get there, and where to go next. Other works extend these definitions of effective feedback by including elements such as motivational and metacognitive aspects \cite{nicol_formative_2006} or aspects of teaching (\eg lesson design; \citealp{evans_making_2013}). 

\vspace{-4pt}
\paragraph{Learner Regulation} \textit{Effective feedback produces a positive response in the learner.} \citet{kluger_effects_1996} argue that, in response to feedback, a learner's attention will be directed to one of three levels: how to solve the task, the task as a whole, or meta-task processes (processes the learner performs while doing the task). Other works note that effective feedback also enhances self-regulated learning behaviors \cite{nicol_formative_2006,evans_making_2013}. \citet{narciss_how_2004, narciss_feedback_2008} extend this definition by arguing that feedback can have three distinct types of impact: influence on the learner's cognitive abilities, their metacognitive skills, or their motivation and self-regulation. \citet{anastasiya_a_lipnevich_david_a_g_berg_jeffrey_k_smith_toward_2016} defend that when a student receives feedback, they produce cognitive and affective responses, judging the task, their level of control, and the feedback. This produces a behavioral reaction, influencing their performance and learning. Similarly, \citet{panadero_review_2022} state that feedback impacts both the students' performance and learning as well as their affective processes and self-regulation.

\vspace{-4pt}
\paragraph{Personalization} \textit{Different types of feedback are best suited for different learner characteristics} \citep{mason_providing_2001}. The learner's individual characteristics will also directly impact how they process feedback \citep{anastasiya_a_lipnevich_david_a_g_berg_jeffrey_k_smith_toward_2016}.

\subsection{What are the characteristics of feedback?}
\label{ss:feedback_characteristics}

A large body of research has attempted to systematically categorize feedback based on its \textit{content}, how it is given (\textit{timing}, \textit{source}), and the variables influencing it (\textit{task}, \textit{learner}).
Works that systematically categorize feedback can be broadly divided into two groups: taxonomies focusing only on the content of feedback, and taxonomies taking into account the whole ecosystem.\iftoggle{appendixes}{\footnote{Appendix \ref{app:feedback_definitions} presents a more thorough definition of proposed feedback categories in the learning sciences.}}{}
\vspace{-4pt}
\paragraph{Content} 
Works in this category focus on the characteristics of the content only. \citet{kulhavy_feedback_1989}, for example, model feedback through a \textit{verification} component, which is a simple discrete classification of the answer as correct or incorrect, and an \textit{elaboration} component, which contains all other information. Other works \cite{hattie_power_2007, panadero_review_2022} suggest three categories for classifying feedback: (i) addressing the learner's performance goal, (ii) addressing the learner's current performance, and (iii) addressing the next steps the learner should undertake. Our feedback content taxonomy draws primarily from the analysis and adaptation of these works to the NLP domain. Additionally, several categorizations include feedback about mistakes, from which we derive the Error component of \ours (§\ref{sec:framework}).
\vspace{-4pt}
\paragraph{Ecosystem} Other works suggest a more comprehensive categorization of feedback including the whole feedback ecosystem. For example, several works propose different feedback categories that consider characteristics of the learner (student proficiency, prior knowledge) and the task (difficulty) \citep{mason_providing_2001,narciss_how_2004,narciss_feedback_2008}. We further explore this categorization in §\ref{sec:feedback_vars}.

\subsubsection{How is feedback given?}
\label{sec:feedback_how}
Apart from its content and ecosystem, feedback is also characterized by how it is given. Two main components have emerged in the literature: 

\vspace{-4pt}
\paragraph{Source} Feedback can be given by different sources (\eg teachers, peers, or even the learner themself). In their systematic review, \citet{lipnevich_review_2021} found that seven out of the 15 considered models view the source as an important characteristic of feedback. An additional three works distinguish feedback generated by an external source from self-feedback.

\vspace{-4pt}
\paragraph{Timing} Previous work has differentiated between immediate and delayed feedback. While early work \cite{bangert-drowns_instructional_1991} found delayed feedback to be more effective, more recent works \cite{mason_providing_2001} argue that the optimal timing of feedback depends on learner characteristics. For example, \citet{hattie_power_2007} state that the most beneficial timing depends mainly on the complexity of the task. Complex tasks benefit from delayed feedback as they allow the learner to properly process the task. Other works \cite{narciss_feedback_2008} focus mainly on the learner characteristics, arguing that as long as the learner possesses the metacognitive skills required to spot and address mistakes, feedback should be delayed.

\noindent We incorporate both feedback source and timing in our framework presented in §\ref{sec:framework}.\footnote{Other points of contention exist, such as feedback valence, but we consider valence as an element of the feedback's content, and as such discuss it only in §\ref{sec:taxonomy}.}

\subsubsection{What variables influence feedback?}
\label{sec:feedback_vars}
Feedback cannot only be categorized according to its content and the way it is given, but is also characterized by the ecosystem surrounding it. Many authors mention the challenge of determining the optimal feedback type in isolation. Instead, characteristics of the \textit{task} and \textit{learner} must be taken into account when giving feedback. We discuss their pedagogical definitions in this section, and subsequently adapt both \textit{Task} and \textit{Learner} specifically to LLMs in our framework, \ours (§\ref{sec:framework}).

\vspace{-4pt}
\paragraph{Task} The task's characteristics influence the optimal timing of feedback (§\ref{sec:feedback_how}) as well as its content. \citet{mason_providing_2001} take into account the complexity of the task when choosing the most suitable feedback for a given setting and \citet{narciss_how_2004, narciss_feedback_2008} incorporate the task and the instructional content and goals into the instructional factors that affect feedback. The nature of the task (\eg closed vs. open-answer) also influences feedback content and processing \cite{lipnevich_review_2021}.

\vspace{-4pt}
\paragraph{Learner} 
\citet{mason_providing_2001} consider student achievement and prior knowledge as important factors impacting feedback optimality. 
\citet{nicol_formative_2006} expand prior knowledge and proficiency into \textit{domain knowledge} and \textit{strategy knowledge} (along with \textit{motivational beliefs}), updated upon each attempt through both internal and external feedback. 
\citet{narciss_how_2004,narciss_feedback_2008} flesh out the learner characteristics further, including learning goals and motivation. \citet{anastasiya_a_lipnevich_david_a_g_berg_jeffrey_k_smith_toward_2016} identify ``personality, general cognitive ability, receptivity to feedback, prior knowledge, and motivation'' as key learner characteristics that impact feedback processing.

Other works focus on the learner's feedback processing mechanisms. \citet{kluger_effects_1996} propose three processing levels: details about how to solve the task, the task itself as a whole, and meta-task processes.
In contrast, \citet{hattie_power_2007} present four levels of feedback processing: (1) \textit{task level}, how well the tasks are understood and performed, (2) \textit{process level}, the process needed to achieve the \textit{task level} understanding and performance, (3) \textit{self-regulation level}, the self-monitoring and the direction and regulation of the learner's actions, and (4) \textit{self level}, personal evaluations about the learner as a whole {(which are usually uninformative about the task and can be very hard to change, \eg the quality of being ``a good student'')}.
Current LLMs do not seem capable of more metacognitive levels of feedback processing.

\section{\ours: Unifying The Two Worlds}
\label{sec:framework}

\begin{figure}[t]
\centering
\includegraphics[width=0.95\columnwidth]{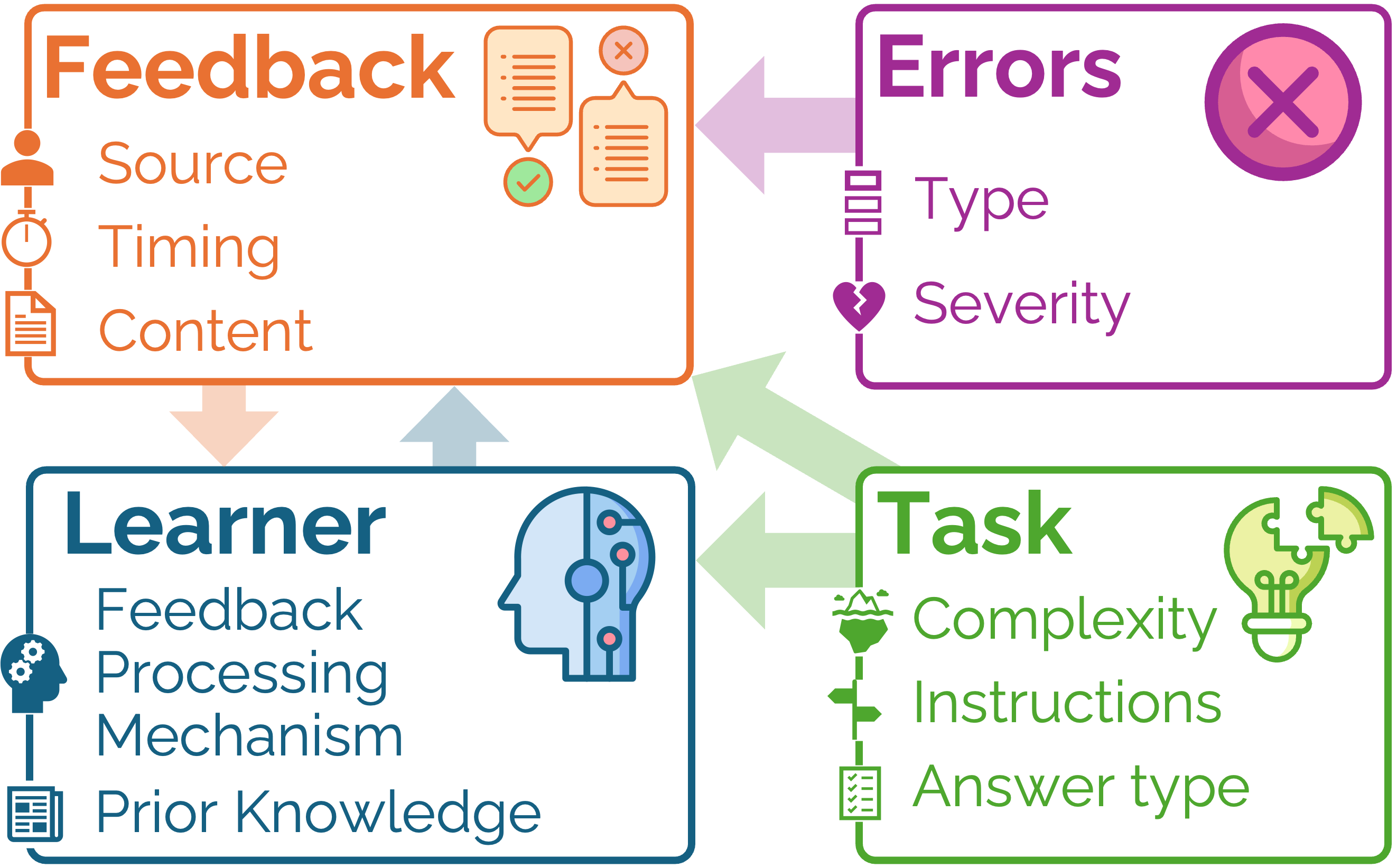}
\caption{\textbf{FELT}, the feedback ecosystem adapted for LLMs. The feedback's characteristics, the errors, the task, and the learner all influence both the feedback and its subsequent processing by the learner. Various interactions occur between the task, the learner, and the feedback, such as the task complexity and prior knowledge of the learner affecting the timing of the feedback. 
\iftoggle{appendixes}{Appendix \ref{app:felt_interactions} presents a more comprehensive overview of the interactions present in the framework.}{}
}
\label{fig:felt_framework}
\end{figure}

In the pedagogical research surveyed in the previous section (\S\ref{sec:background_pedagogy}), feedback emerges as both a complex ecosystem, and a rich, but systematized, information source. 
In this section, we consolidate different pedagogical feedback models, and conceptualize a novel feedback framework adapted for LLMs, \ours (Feedback, Errors, Learner, Task), displayed in Figure \ref{fig:felt_framework}. We briefly analyze each of its components, and dedicate a specific section (§\ref{sec:taxonomy}) to the formalization of feedback content.

\subsection{The Big Picture}
By consolidating the dimensions presented in Section \ref{sec:background_pedagogy}, we developed FELT, which incorporates the content and delivery of feedback, as well as the rest of the ecosystem: the task, learner, and errors.
\vspace{-4pt}
\paragraph{Task}
The task reflects the work assigned to the LLM. It has three attributes: \textit{complexity} (\ie difficulty level), \textit{instructions}\footnote{Instructions can pertain to a description of the task, to specific criteria the model must satisfy, or to further directions, e.g., how much the model must adhere to any given feedback.} given, and the task's \textit{answer type} --- which we reduce to being closed-answer (where there is a single correct answer or a finite set of them) or open-ended. 
When ported to LLMs, the task component can be reduced to data, as the training data will reflect the distribution of tasks that must be learned. Unsurprisingly, the impact of alignment data on model behavior is an active area of research in the NLP community \citep{eli5_lfqa, pmlr-v162-ethayarajh22a, wang2023selfinstruct, alpaca, guo-etal-2023-hc3, h4stackexchange, köpf2023openassistant, bai2022training}.

\vspace{-4pt}
\paragraph{Learner}
The learner in \ours represents the LLM itself, represented by two components: \textit{prior knowledge} (which is dependent on the task), and the \textit{feedback processing mechanism}.\footnote{{Several pedagogical dimensions for human feedback processing, discussed in §\ref{sec:background_pedagogy}, are not currently applicable to LLMs --- such as the \textit{learner's regulation}, the learner's \textit{affective processes} and multiple \textit{motivations} (\eg both succeeding at the task and being seen as competent, which can be a deterrent to success if the learner believes they might fail), the learner's perception of their \textit{control over the task}, autonomous \textit{receptivity to feedback}, and higher levels of feedback processing, including \textit{meta-task} and \textit{self level} learning --- and as such do not have a direct representation in \ours's current form.}}
\textit{Prior knowledge} is captured by the model's size, pretraining data and pretraining method. Model size is suggested to be linked to the model's ability to effectively leverage feedback \citep{scheurer2022training, bai_constitutional_2022}. The pretraining data, as well as how the model was trained, similarly encode its initial ability to tackle a given task. 
The \textit{feedback processing mechanism} reflects the method used to transmit feedback to the model, naturally influencing how the feedback is subsequently processed. For example, training objectives necessarily influence how models process and incorporate feedback. \citet{fernandes_bridging_2023} identify three common feedback integration mechanisms: feedback-based imitation learning, joint-feedback modeling, and reinforcement learning (which can be generally extended to include other recent training methods, such as DPO, \citealp{rafailov2023direct}). In addition, we also consider feedback provided using in-context learning (ICL; \citealp{brown2020language, madaan-etal-2022-memory,zhao2024incontext}). Much like the \textit{Task}, the \textit{Learner} (i.e., the models and algorithms used for learning from feedback) is an active area of NLP research.

\vspace{-4pt}
\paragraph{Errors}
Feedback may relay information on where the learner is failing, which requires understanding the possible error modes for a given task, and which ones the learner is likely in. For example, guessing and committing systematic reasoning mistakes reflect differing degrees of understanding. Moreover, not all errors should be treated the same. \citet{wu2023finegrained} find that assigning distinct reward models to specific error modes (\eg conciseness, factuality, relevance) improved performance over using a single RM, clarifying the importance of modeling the error space -- another active area of research \citep{golovneva2023roscoe, paul2023refiner, murugesan2023mismatch, mishra2024finegrained}.

\vspace{-4pt}
\paragraph{Feedback}
Pedagogy distinguishes three important variables for feedback: content, source, and timing. 
The \textit{content} captures both the type and the format of the information provided in a feedback message, two aspects we will explore fully in Section \ref{sec:taxonomy}.
The \textit{source} of feedback typically identifies whether the feedback stems from an authority figure or a peer. In the context of LLMs, the source could be either a human or a separate AI model and receiving this information might lead LLMs to behave differently. However, current prevailing LLM architectures can only perceive the source of information by having it conveyed as text input to the model, reducing it to a dimension of content.

\textit{Timing} specifies whether feedback is provided immediately or after a delay. Certain types of prompting can be formulated as variations in timing. For example, Metacognitive reasoning \citep{wang2023metacognitive} prompts the model to answer a question, critically evaluate its answer, possibly revise it, and assess its confidence in it. In contrast, Chain-of-Thought (CoT; \citealp{wei_chain_2022}) does not elicit any post-answer reasoning. Coupled with feedback learning methods such as RLF or DPO, this ``padding'' between the answer and the reward signal might affect which prompting strategy leads to the best performance.
Delayed feedback also has the potential for deriving multiple model updates from a single initial data point. From an answer and a reflection generated by an LLM (\ie an attempt to identify and correct errors on that answer), and a piece of feedback, a model can learn not only whether the answer is correct, but also whether its reflection is adequate --- applicable both in ICL and traditional training (\eg RLF, DPO) settings.

\subsection{Zooming in on Feedback Content}
\label{sec:taxonomy}

Of the four dimensions in \ours{}, feedback emerges as the least systematized, particularly when it is given through text or natural language.
As briefly mention in Section \ref{ss:feedback_characteristics} and detailed in Appendix \ref{app:feedback_definitions}, several works have sought out to characterize feedback focusing only on its content. We draw from this research to define a novel taxonomy for feedback content, for which all dimensions can be controlled for and adjusted by the feedback provider.

\vspace{-4pt}
\paragraph{Defining feedback content}
\label{ssec:feedback_content_what}

We distill four non-overlapping areas from categorizations of feedback in pedagogy research (\S\ref{sec:background_pedagogy}):
\begin{enumerate}[topsep=1ex,itemsep=-1ex,partopsep=1ex,parsep=1ex]
    \item \textbf{Learner status.} Situates the learner's performance. May indicate what the learner got right, what mistakes the learner made, or both.
    \item \textbf{Goal.} Demonstrates the target performance for the task by providing either the correct answer or an example solution.
    \item \textbf{Procedural.} Provides instructions for the learner to follow on a subsequent attempt. These recommendations can be specific to the task, be general problem-solving and metacognitive instructions, or both.
    \item \textbf{Peripheral.} Supplementary information not directly related to the above three areas. Peripheral feedback can include the clarification of the task (without any instructions), or the elaboration of concepts relevant to the task.
\end{enumerate}

\paragraph{Modulating feedback content}
\label{ssec:feedback_content_how}

Once the decision of \textit{what} content to include or omit in the feedback has been taken, \textit{how} the feedback will be given must also be decided. As such, by adapting pedagogical models as well as incorporating entirely new axes exclusive to LLMs, we propose 10 dimensions that capture the form feedback takes, and that can be controlled by the feedback giver:

\begin{enumerate}[itemsep=0.05em]
    \item \colorbox{purple!30}{\textit{Granularity}}: measure of detail with which the feedback addresses the original answer.\footnote{For an open-answer example task, feedback might range from global meta-feedback, to task-specific, to paragraph-level, to sentence-level, to word-level, to token-level feedback. Appendix \ref{app:feedback_examples} provides several examples.} 
    \item \colorbox{teal!30}{\textit{Applicability of instructions}}: outlines if the feedback contains instructions, and how applicable those instructions are for the learner and their current approach to solving the task.
    \item \colorbox{magenta!30}{\textit{Answer coverage}}: registers how much of the learner's answer is reflected in the given feedback. The feedback could be independent of the answer, relate only to parts of the answer (\eg focusing on a particular mistake), or take the complete answer into consideration.
    \item \colorbox{orange!40}{\textit{Target coverage}}: indicates how much of the target performance is being considered when generating the feedback. \textit{Goal} and \textit{Procedural} types of feedback will both have at least some degree of \textit{target coverage}.
    \item \colorbox{blue!30}{\textit{Criteria}}: denotes which criteria the answer is being evaluated on: global evaluation, specific dimensions (e.g., fluency, engagement, etc.), or, alternatively, no  dimensions (the answer is not being evaluated).
    \item \colorbox{green!20}{\textit{Information novelty}}: indicates the degree to which learner already had access to the information in the feedback, ranging from all information being previously known, to some information being unknown to the learner, to all information being novel for the learner.
    \item \colorbox{brown!40}{\textit{Purpose}}: measures whether the feedback is being given to improve the learner's performance or to clarify the task.\footnote{In a pedagogical setting with human learners, other purposes are possible, such as regulating the student's emotions and motivation, but we do not consider these for LLMs.}
    \item \colorbox{cyan!30}{\textit{Style}}: captures the type of language used to transmit the feedback to the learner, which can range from simple, direct sentences to verbose and terminology-heavy language.
    \item \colorbox{olive!30}{\textit{Valence}}: indicates whether the feedback is positive (signaling achievement) or negative (signaling need for improvement).
    \item \colorbox{red!30}{\textit{Mode}}: captures
    whether the feedback is uni- or multimodal.\footnote{Multimodal feedback is naturally more suited for multimodal tasks. For example, in an instance segmentation task, the correct (visual) answer could be provided alongside textual feedback on mistakes and how to correct them.}
\end{enumerate}

\noindent The ability to modulate these ten dimensions enables practitioners to specifically craft feedback for their use cases and learning frameworks. 
Moreover, our feedback taxonomy is general and transferable across domains. While tasks can vary significantly in both type (\eg coding, mathematics, story telling, common sense) and necessary skills (\eg reasoning, planning, retrieval), the same type of feedback applies to all of them.

\section{Realizing the Promise of Feedback}
\label{sec:feedback_applications}

\begin{figure*}[t]
\centering
\includegraphics[width=0.8\textwidth]{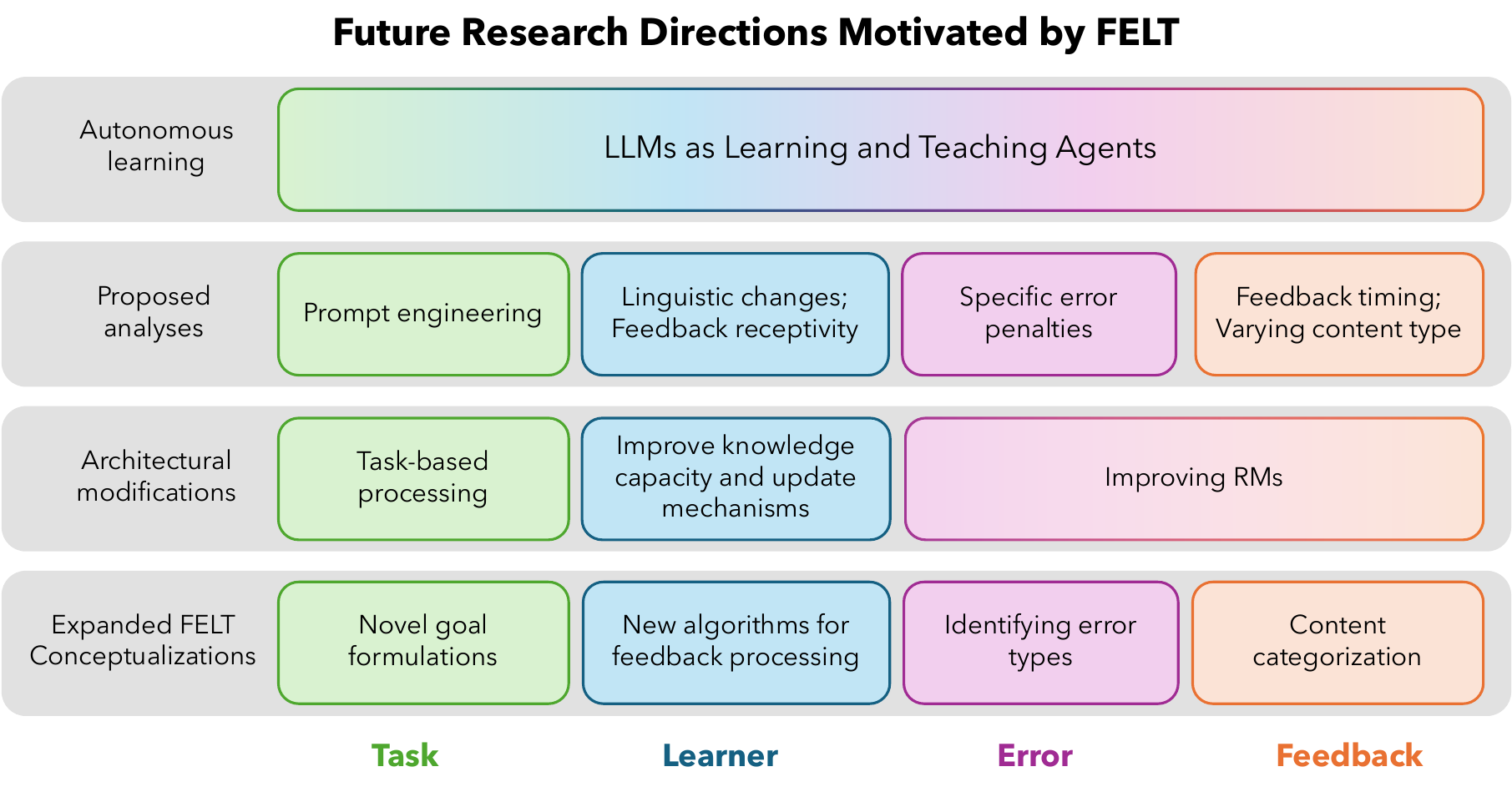}
\caption{{\textbf{Visual summary of several future research directions motivated by FELT,} for each of its dimensions and across several research axes. The textual counterparts of each bubble are highlighted in their respective color (\textcolor{task_green}{\textbf{green}}, \textcolor{learner_blue}{\textbf{blue}}, \textcolor{error_purple}{\textbf{purple}}, \textcolor{feedback_orange}{\textbf{orange}}) in the main body of Section~\ref{sec:feedback_applications}.
}}
\label{fig:felt_future}
\end{figure*}

\ours provides a springboard for exploring the entire feedback ecosystem. To understand what makes feedback effective for LLMs, we briefly propose some ideas for future investigation (and categorize ideas from previous studies) that demonstrate the important role that each of the different components of \ours play in this exploration. {Figure \ref{fig:felt_future} presents a visual summary of the proposals presented in this section.}
    
    \noindent \textit{Task.} Feedback might influence model performance differently depending on the \textbf{\textcolor{task_green}{goal formulation}} for the LLM (\eg asking the model to extract entities from text v.s. asking the model to identify their start and end positions). {\textbf{\textcolor{task_green}{Task-specific processing}} (\eg LoRA; \citealp{hu2022lora}) would allow models to adapt to a wider range of tasks.}
    
    \noindent \textit{Learner.} To explore the degree to which alignment extends beyond \textbf{\textcolor{learner_blue}{changing the linguistic style}} of AI models \citep{lin2023unlocking}, {LLMs can be modeled as different types of learners --- whether by (1) controlling their \textbf{\textcolor{learner_blue}{knowledge capacity}}, \ie creating learners both with and without prior knowledge through data interventions during pretraining or finetuning, or external augmentation; and (2) implementing \textbf{\textcolor{learner_blue}{novel update mechanisms}} that can capture richer feedback (\eg TextGrad; \citealp{yuksekgonul2024textgrad}).} 
    Uncertainty awareness has also been show to improve model alignment \citep{wang2024uncertainty}.
    Exploring the \textbf{\textcolor{learner_blue}{model's receptiveness to feedback}} conditioned on its uncertainty may enable better understanding of what leads models to reject their own parametric knowledge. 
    
    \noindent \textit{Errors.} {After \textbf{\textcolor{error_purple}{mapping the different error types}} a model might make, it becomes possible to implement \textbf{\textcolor{error_purple}{different penalties for these error types}}, which might promote different model behaviors \citep{wu2023finegrained}.} This flexibility will yield a better understanding of which degrees of penalization lead to a given behavior, as well as the degree to which this penalization should be aligned with the actual severity that the error merits.
    
    \noindent \textit{Feedback.} By delaying the \textbf{\textcolor{feedback_orange}{timing of the feedback}} and first asking the model for a reflection, it is possible to multiply the datapoints the model is trained on (generating feedback not only on the correctness of the answer but also on that of the model's reflections). This approach has, to the best of our knowledge, not yet been pursued.

\vspace{-4pt}
\paragraph{\textbf{\textcolor{feedback_orange}{Categorizing Feedback}}} A limitation of prior works on NLF is the indiscriminate treatment of different types of feedback ({shown in Table \ref{tab:previous_works}}). The four types of feedback content --- \textit{learner status}, \textit{goal}, \textit{procedural}, and \textit{peripheral} --- directly address this issue, allowing more systematic studies of feedback. The natural next step is to provide models with \textbf{\textcolor{feedback_orange}{various kinds of feedback}} and compare the impact of the different types of information and of its delivery.

\begin{table*}[t]
 \centering
 \fontsize{9.0pt}{\baselineskip}\selectfont 
 \begin{tabular}{l l}
 \toprule
\textbf{Content Type} & \textbf{Previous Works} \\
\midrule \midrule

\textbf{Learner Status} & \parbox{13cm}{\citet{weston_dialog-based_2016}, \citet{tandon-etal-2022-learning}, \citet{scheurer2022training}, \citet{saunders2022selfcritiquing}, \citet{shi2022life}, \citet{welleck2022generating}, \citet{wu2023finegrained}, \citet{shinn2023reflexion}, \citet{chen2023teaching}, \citet{paul2023refiner}, \citet{madaan2023selfrefine}, \citet{lightman2023lets},
\citet{yoon-etal-2024-tlcr}}
\\
\midrule
\textbf{Goal} & \parbox{13cm}{\citet{weston_dialog-based_2016}, \citet{saunders2022selfcritiquing}, \citet{shi2022life}}
\\
\midrule
\textbf{Procedural} & \parbox{13cm}{\citet{weston_dialog-based_2016}, \citet{tandon-etal-2022-learning}, \citet{saunders2022selfcritiquing}, \citet{shi2022life}, \citet{welleck2022generating}, \citet{madaan2023selfrefine}, \citet{murty2022fixing}, \citet{schick2022peer}}
\\
\midrule
\textbf{Peripheral} & \parbox{13cm}{\citet{tandon-etal-2022-learning}, \citet{scheurer2022training}, \citet{chen2023teaching}, \citet{madaan2023selfrefine}}
\\
 \bottomrule
 \end{tabular}
 \caption{{
 \textbf{Past works using textual or otherwise augmented feedback.} 
Even in the same work, feedback such as ``write a step-by-step reasoning trace before the solution'' is often treated as equivalent to ``should be contextually relevant and easy to check.'' As made evident by our taxonomy, however, they are markedly different, as the former provides \textit{procedural} instructions and the latter \textit{peripheral} information clarifying the task.
 }}
 \label{tab:previous_works}
 \vspace{-4mm}
 \end{table*} 

\vspace{-4pt}
\paragraph{\textbf{\textcolor{task_green}{Prompt Engineering}}} ICL is an active area of research in NLP, including in model alignment
\citep{lin2023unlocking, zhao2024incontext}. Several dimensions of our feedback content taxonomy have been shown to contribute to either aligning LLMs to specific desired behaviors, or to increasing its performance in a given task --- suggesting both the utility of these content dimensions, and hinting at many as of yet unexplored approaches that vary them in novel ways. 
Prompts with a high degree of \colorbox{teal!30}{\textit{applicability of instructions}} and with a focus on \textit{procedural} types of information are common, be it asking for a rationale 
\citep{wei_chain_2022, wang2023metacognitive, yao2023tree}, 
or more task-specific instructions 
\citep{wang2023metacognitive, madaan2023selfrefine}. Prompting the model with \textit{goal} information is also popular, most frequently done using few-shot prompting \citep{brown2020language}.

\colorbox{olive!30}{\textit{Valence}} is often introduced either from (explicit or implicit) human feedback \citep{shi2022life, pang2024leveraging} or by criticizing a preliminary answer \citep{bai_constitutional_2022, paul2023refiner, yao2023tree, yuan2024selfrewarding}, providing the model with information on its \textit{learner status}.
The \colorbox{cyan!30}{\textit{style}}, or linguistic properties, of prompts can lead to significant performance variations \citep{leidinger2023language}.
Finally, while \colorbox{green!20}{\textit{Information Novelty}} is hard to measure for LLMs, recent research has investigated what causes LLMs to be faithful to the novel information rather than their encoded knowledge \citep{Longpre2021EntityBasedKC}, an important quality to understand and predict model behavior \citep{zhou2023contextfaithful, yin2023alcuna}.
All these dimensions underlie a wide range of instruction formulations for LLMs, enabling practitioners to taxonomize existing prompting strategies, and uncover novel approaches based on different combinations.

\vspace{-4pt}
\paragraph{\textbf{\textcolor{error_purple}{Improving Rew}\textcolor{feedback_orange}{ard Models}}}
Textual feedback can be integrated into the training loop of reward models, whether in a traditional training pipeline (\ie RLF) or in inference-time learning (\eg Inference-Time Policy Adapters; \citealp{lu2023inferencetime}). For example, textual feedback about a policy model's output could be provided in addition to its output to the reward model, or, following \citet{wu2023finegrained}, the model can be rewarded according to a more targeted set of preferences. It is possible to go further still by varying the \colorbox{purple!30}{\textit{granularity}} of the feedback, allowing targeted rewards for different parts of the model response to be generated. %
A similar effect can by achieved by considering a set of different \colorbox{blue!30}{\textit{criteria}} on which to judge model performance. RMs themselves can be extended, by adding adapters for different criteria or using a Mixture-of-Experts (MoE) to better model these fine-grained preferences.
The exploration of these dimensions allows for more expressive RMs, and possibly modeling non-deterministic preferences. 

\vspace{-4pt}
\paragraph{\textbf{\textcolor{task_green}{LLMs as }\textcolor{learner_blue}{Learning a}\textcolor{error_purple}{nd Teachin}\textcolor{feedback_orange}{g Agents}}}
By defining the various factors that impact learning from feedback, \ours allows the creation of a \textit{environments} to explore and optimize feedback for LLMs. Multiple LLM agent learners can be simulated in this environment, each receiving differently formulated feedback to guide them toward their goals. This environment could enable the study of the learning process from initial exposure to data to more complex interaction with other learner agents. 
The mapping provided by \ours similarly enables the deployment of LLMs as teachers by identifying the necessary components needed for better feedback dataset construction, making feedback more adapted and personalized for learner models.
\section{Conclusion}
We survey the most influential feedback models from the learning sciences, creating a novel framework, \ours, as well as a taxonomy of natural language feedback (NLF). Both enable the systematic exploration of feedback, allowing for objective conclusions on the impact of a piece of feedback and the optimization of NLF.
Beyond a survey of the space of natural language feedback, we explore how both \ours{} and the fine-grained feedback content dimensions underlie many existing techniques in natural language feedback in AI, making specific recommendations for enhancing feedback formulations for a wide range of proposed topics.

\section*{Limitations}
\ours, our proposed framework to capture the full feedback ecosystem, is only theoretically grounded at the moment. Certain aspects of our taxonomy, such as the impact of feedback timing, need to be empirically assessed for LLMs. Similarly, while we conjecture that all 10 dimensions of the feedback content taxonomy will impact how models react to feedback, this inference has not yet been empirically confirmed. However, our grounding of the LLM feedback space to pedagogical principles is meant to  provide a broad framework for organizing feedback research, with each component available for empirical validation by the research community. Components of our framework that end up failing empirical validation perhaps indicate areas where LLMs differ from human learning. 

Another limitation of our work is that both pedagogical and NLP conceptualizations and results we discussed in this paper were conducted in English settings. While we expect \ours and our feedback content taxonomy to generalize to other languages, the same feedback content might affect models differently depending on the language with which it is given. This dimension must be taken into account by future work. Finally, any study into how to make feedback more effective has the potential to contribute to the jailbreaking of LLMs or other purposefully malicious changes in its behavior. However, we also not that better understanding the components of feedback that make it effective will enable researchers to develop models that are better aligned to their original goals and perhaps more robust to these types of attacks.

\section*{Acknowledgements}
We gratefully acknowledge the support of the Swiss National Science Foundation (No. 215390), Innosuisse (PFFS-21-29), the EPFL Science Seed Fund, the EPFL Center for Imaging, Sony Group Corporation, and the Allen Institute for AI.

\bibliography{custom,anthology1,anthology2}

\iftoggle{appendixes}{
\appendix
\section{Components of the FELT Framework}
\label{app:felt_interactions}

The FELT framework introduced in Section \ref{sec:framework} presents an important overview of all the factors that influence feedback and are in turn influenced by it. Figure \ref{fig:felt_framework} showcased a schematic overview of the FELT framework, integrating four distinct components: Feedback, Errors, Learner, and Task. In this appendix, we will outline more precisely each of the components of the FELT Framework, as well as the interactions between them.

\subsection{Task}
Typically, the task will be the first element to be defined. 

\paragraph{Answer Type} 
Understanding the answer is fairly straightforward -- a task has a closed-answer if there is a finite set of correct answers, and an open-answer otherwise. Notably, tasks can contain both elements. For example the task "\textit{Write a quality 4-paragraph short-story}" has both open- and closed-answer elements. There is no finite set of answer of what a quality story is, but whether a story has 4 paragraphs, or not, is a binary closed-answer task.

\paragraph{Complexity} The difficulty level of a task is harder to define as some measure of relativity is involved. We suggest anchoring this measurement to the average adult human capabilities. A simple arithmetic task will thus be considered very easy, whereas researching and writing a doctoral thesis would be seen as hard.

\paragraph{Prompt Instructions} The task instructions will be presented to the model at two distinct points in time: when first assigning the model this task, and when later providing feedback. With regards to the former, this element captures the degree to which the task is explained – is the model explicitly aware of all criteria it should satisfy? With regards to the second pass, when feedback is provided, this dimension pertains instead with the degree of freedom it gives the LLM – is the model forced to take the feedback into account, or can it consider only part of it, or even disregard it altogether if it deems it useless?

\subsection{Learner}
Either at the same time the task is defined or immediately after, the model to be tested will be chosen. The model choice influences two important features.

\paragraph{Prior Knowledge} The prior knowledge captures the LLM's abilities as a direct result of its size, training data, and training method. These, in turn, also reflect the model's purpose (e.g., was it designed to be helpful, harmless, entertaining, etc.). The prior knowledge thus captures the model's  representation of the learner, and in its architecture and parametric knowledge, it encodes the LLM's current abilities -- or its proficiency -- both in general and with regards to the specific task.

\paragraph{Feedback Processing Mechanism} Mainly defined by the experimental setup, the mechanism by which the model process feedback can vary significantly, and not all of them are able to leverage the same level of information. Imitation learning, for example, can only leverage information which was positively evaluated. As stated in Section \ref{sec:framework}, we identify 4 main processing mechanisms, 3 of which alter the model's parametric state -- feedback-based imitation learning, joint-feedback modeling, and reinforcement learning, as defined in \citet{fernandes_bridging_2023} -- and a fourth, non-parametric mode: in-context learning \citep{brown2020language}.

\subsection{Errors}
After both the task and learner are in place, the first pass of the experiment can be run, where the model will have its first attempt at solving the task. In this attempt, it is expected that the model will make some degree of mistakes -- which have two important characteristics.

\paragraph{Error Type} There are several possible types of errors, and their differences are significant. For example, an error made due to a guess only needs to provide the learner with the right information for it be be corrected, whereas a systematic error (for example, the mixing of British and American English spellings) will require a different, much more insistent, intervention. ROSCOE \citep{golovneva2023roscoe} proposes a taxonomy of step-by-step reasoning errors. While task dependent (i.e., there are grammar errors and arithmetic errors, rather than fully task independent failure modes), this taxonomy provides a good starting ground for the exploration of error types in NLP.

\paragraph{Error Severity} Besides the type of error, it is also important to take the severity of the error into account. Stating that Marie Curie was a German philosopher and stating that she won one Nobel Prize in her lifetime are both factually inaccurate -- but one is a severe, complete hallucination, while the other omitted she actually won the Nobel Prize twice. The more severe the error, the stronger, more insistent, and more corrective the feedback should be.

\subsection{Feedback}
Finally, after the model has finished its first attempt at the task, producing some number of errors, feedback can be provided on this attempt.

\paragraph{Timing} One easy to neglect aspect of feedback that pedagogy has shown to be impactful is timing -- whether the feedback is provided immediately after a task is attempted or whether there is a delay between the two actions. There are differing opinions amongst education researchers, but how to make feedback content more effective through timing merit research in LLMs. For example, in line with \citet{Mathan2005FosteringTI} and \citet{narciss_feedback_2008}'s take on timing -- delay feedback if the learner possesses metacognitive abilities that allow them to identify and possibly correct mistakes -- we posit feedback will be more effective if, content-wise, it is preceded by information on the answer's correctness and mistakes' existence and only after this metacognitive priming is the rest of the information presented.

\paragraph{Content} Section \ref{sec:taxonomy} explores feedback content in depth, presenting 10 impactful axes on which it can vary: length, granularity, applicability of instructions, answer coverage, criteria, information novelty, purpose, style, valence, and mode. It also presents a set of 9 emergent categories which, based on pedagogical research, we estimate to be the most promising one with regards to impact on revised model generations, and thus most deserving of further study.

\paragraph{Source} Finally, it is also important to consider the source of feedback, which might be an authority, such as an expert, an average human, another LLM, a rule-based system, among others. Different sources will reflect different authority and reliability levels.

\subsection{Interactions}
With a clear understanding of all the components and sub-components of the FELT framework, we can explore the influences that exist between them.

Both the task complexity and the learner's prior knowledge can impact the ideal feedback timing -- be it delayed when the learner has metacognitive skills \citep{narciss_feedback_2008} or enough task proficiency \citep{mason_providing_2001} they can identify where the mistake occurred, or, for example, immediate if they don't \citep{narciss_feedback_2008} or the task difficulty is low \citep{mason_providing_2001}.

With regards to the feedback content, the type of task \citep{butler_feedback_1995, kluger_effects_1996, mason_providing_2001, anastasiya_a_lipnevich_david_a_g_berg_jeffrey_k_smith_toward_2016} and both the error type and severity will have an impact \citep{narciss_how_2004, narciss_feedback_2008}. The nature of the task (open or closed answer) will directly condition the feedback that can be given in response to the model's answer, as well as how difficult it will be to produce it. For example, generating the correct answer for a multiple choice quiz or a story writing task will be two very different endeavors. Similarly, it is impossible to provide response elaboration feedback on a single multiple choice question.
The error type and severity will also influence the feedback content, as apart from directly dictating what mistakes verification and elaboration feedback can be given, they will also condition the ideal amount of detail and explanations to address the mistake at the most efficient level.

Finally, all aspects of feedback will influence the learner's feedback processing mechanism \citep{kulhavy_feedback_1989, sadler_formative_1989, bangert-drowns_instructional_1991, butler_feedback_1995, kluger_effects_1996, narciss_how_2004, nicol_formative_2006, narciss_feedback_2008, anastasiya_a_lipnevich_david_a_g_berg_jeffrey_k_smith_toward_2016, carless_development_2018}. All three dimensions of feedback have evident potential to directly influence how the model processes them. The instruction's permissiveness to consider or discard feedback will also impact the learner's feedback processing mechanism. This processing is, of course, dependent on the specific processing mechanism employed, and while some might be indifferent to some of these components -- like imitation learning, for example, which focuses exclusively on the feedback content -- others will be sensitive to all, including the task's prompt instructions -- such as in-context learning.
\section{Feedback content examples}
\label{app:feedback_examples}

To help concretize the 10 dimensions alongside which feedback content can be modulated, in this appendix we provide a few examples for each of these dimensions, with the exception of \textit{mode}, which simply presents information in formats beyond text (\eg images, audio, video).

To achieve this, we will consider the scenario below, considering two different model answers to better showcase different feedback formulations. In reality, \colorbox{blue!30}{Model Answer B} would likely be a revised version of \colorbox{pink}{Model Answer A} after feedback along the lines of that presented in \ref{app:ssec:novelty}.

\paragraph{\colorbox{yellow}{Initial Prompt}}
\texttt{Please provide me with general information on the European Parliament Elections that took place on June 9, 2024.}
\paragraph{\colorbox{pink}{Model Answer A}}
\texttt{I cannot provide an answer as this event takes place after my training data cutoff date.}

\paragraph{\colorbox{blue!30}{Model Answer B}}
\texttt{More than 20 European countries voted over 720 seats. The European People’s Party is expected to have the most seats out of any party.}

\subsection{Granularity}
\label{app:ssec:granularity}
Below we present some examples of feedback at different levels of granularity for \colorbox{blue!30}{Model Answer B}, with a focus on the \textit{procedural} type of information. Many more levels of grnaularity are possible, and it can also be combined (\eg provide very granular feedback on mistakes, and general feedback on correct parts of the answer).
\begin{itemize}
    \item \textbf{General granularity.} The answer lacks detail --- you should specify the number of countries, the number of seats, and other such details.
    \item \textbf{Sentence granularity.} In the first sentence, you should specify the number of countries (27). You can enumerate all 27 to achieve more clarity. In the second sentence, you should precise the number of seats won by that party, as well as other top parties, their political ideology (left, center, right), whether they can constitute a majority with other parties of the same ideology, etc.
    \item \textbf{Word granularity.} Replace the first three words with 27, the exact number of countries. After 'European countries' add, in parenthesis, a list of the 27 countries names. Add a new sentence detailing the several parties and their political affiliation. After four words in the now third sentence, add their acronym in parentheses and at the end of the sentence add the exact number of seats they have. Add a fourth sentence stating whether they can achieve majority with other parties with the same political ideology.
\end{itemize}

\subsection{Applicability of instructions}
\label{app:ssec:applicabilty}
Below we present some examples of feedback at different levels of instructions' applicability for \colorbox{blue!30}{Model Answer B}, showcasing three well-defined points of this spectrum:
\begin{itemize}
    \item \textbf{Concrete instructions.} Specify the number of voting countries (27), and enumerate them. List all the parties with their political ideology. State whether a majority is possible for the party with the most votes.
    \item \textbf{Metacognitive instructions.} Break the request into several sub-tasks, and enumerate them. Then, answer each sub-task individually. Once you are done, check if the initial question has been fully answered. If not, address any points not yet covered by your answer.
    \item \textbf{No instructions.} Your answer is satisfactory, but it could be better.
\end{itemize}

\subsection{Answer coverage}
\label{app:ssec:answer_cover}
Below we present some examples of feedback at different levels of answer coverage for \colorbox{blue!30}{Model Answer B}. More combinations are possible (\eg covering both mistakes and correct parts for only a subset of the answer, exploring the order in which each part of the answer is covered, etc.).
\begin{itemize}
    \item \textbf{Full answer.} While you provided a concise overview and identified the leading political party, you could have provided more detail --- such as the political ideology of the party you mention, and more concrete details overall (specify it's 27 countries, the number of seats the leading party won out of the 720 total seats, etc.)
    \item \textbf{Successful parts.} You provided a concise overview and identified the leading political party.
    \item \textbf{Lackluster parts.} You could have provided more detail --- such as the political ideology of the party you mention, and more concrete details overall (specify it's 27 countries, the number of seats the leading party won out of the 720 total seats, etc.)
\end{itemize}

\subsection{Target coverage}
\label{app:ssec:target_cover}
Below we present some examples of feedback at different levels of target coverage for \colorbox{blue!30}{Model Answer B}. More combinations can be done.
\begin{itemize}
    \item \textbf{Full target (correct answer).} On June 9, 2024, citizens of the 27 European Union countries voted for the 720 European Parliament seats. EPP, European People’s Party, is expected to win the most seats, 189. 361 seats are needed for a majority, which seems achievable if the three centrist parties come together: EPP (189 seats), S\&D (Progressive Alliance of Socialists and Democrats; 135 seats) and Renew (Renew Europe; 79 seats) --- achieving a total of 403 seats.
    \item \textbf{Correcting lackluster parts.} Rather than more than 20 countries, it was exactly the 27 member states of the European Union who voted for these elections. The European People’s Party is expected to win 189 seats. If the three centrist parties come together (European People’s Party, Progressive Alliance of Socialists and Democrats, Renew Europe), they can achieve a majority at a total of 403 seats.
    \item \textbf{Contrasting satisfactory parts.} You could mention the date just to clarify which elections you are referring to in your answer. You could also have added the acronym for the European People’s Party (EPP).
\end{itemize}

\subsection{Criteria}
\label{app:ssec:criteria}
Below we present some criteria that could be used to evaluate and provide feedback for \colorbox{blue!30}{Model Answer B}:
\begin{itemize}
    \item \textbf{Factuality.} The answer is factually correct.
    \item \textbf{Impartiality.} The answer is impartial and unbiased, providing information without an underlying goal or narrative.
    \item \textbf{Completeness.} The answer is very incomplete, mentioning only one party without exploring the overall distribution of votes nor the parties' political affiliations.
    \item \textbf{Clarity.} The answer is fairly clear and readable, but needlessly vague on some section (not specifying the exact number of European Union member states, not indicating the number of seats won by the European People’s Party).
    \item \textbf{Relevance.} The answer is relevant to, and in the domain as, the question.
\end{itemize}

\subsection{Information novelty}
\label{app:ssec:novelty}
An example of feedback containing novel information for \colorbox{pink}{Model Answer A}, which lacks the necessary parametric knowledge to be able to provide an answer:
\begin{itemize}
    \item \textbf{Novel Information.} On June 9, 2024, citizens of the 27 countries that make up the European Union (Germany, France, Italy, Spain, Poland, Romania, Netherlands, Belgium, Czech Republic, Sweden, Portugal, Greece, Hungary, Austria, Bulgaria, Denmark, Finland, Slovakia, Ireland, Croatia, Lithuania, Slovenia, Latvia, Estonia, Cyprus, Luxembourg, Malta), voted for the 720 European Parliament seats. EPP, European People’s Party, is expected to win the most seats, 189. 361 seats are needed for a majority, which seems achievable if the three centrist parties come together: EPP (189 seats), S\&D (Progressive Alliance of Socialists and Democrats; 135 seats) and Renew (Renew Europe; 79 seats) --- achieving a total of 403 seats.
    \item \textbf{Known Information.} The European Parliament Elections took place on June 9, 2024, and had European countries vote on parties for the European Parliament.
\end{itemize}

\subsection{Purpose}
\label{app:ssec:purpose}
Below we present two contrasting examples of feedback with different purposes for \colorbox{blue!30}{Model Answer B}. Many variations are possible.
\begin{itemize}
    \item \textbf{Improving performance.} Specify the number of voting countries (27), and enumerate them. List all the parties with their political ideology. State whether a majority is possible for the party with the most votes.
    \item \textbf{Clarifying the task (concretely).} You should provide an answer that concisely presents the most important information at the start (number of voting countries, parties who won the most seats, their political ideology and whether a majority can be established). Then in a subsequent paragraph you can list all the parties, their affiliations, and seat count, and well as the 27 countries. If you want to provide an even more complete answer, you can break down the votes per country and analyze the trends you find there.
    \item \textbf{Clarifying the task (peripherally).} To write a good answer, you should read the question carefully and ensure your answer either addresses the question in its entirety or at least part of the question if it is not feasible to answer it in one go. Make sure to adopt a polite tone and strive for a clear and understandable answer. If the question is not clear or not well formulated, start by asking clarifying questions before attempting to answer. If you do not know the answer, honestly admit that.
\end{itemize}

\subsection{Style}
\label{app:ssec:style}
Below we present examples of feedback with different style for \colorbox{blue!30}{Model Answer B}. Many variations are possible, and we expect the model's answer to reflect the style used in some linguistic artifacts (\eg if a particularly informal tone was used, we expect the models' answer to lean toward lower formality as well).
\begin{itemize}
    \item \textbf{Normal.} Please rewrite your answer, but this time specify the number of voting countries (27), list all the parties with their political ideology, and state whether a majority is possible for the party with the most votes.
    \item \textbf{Informal.} How about you try again, but this time make sure to say 27 countries explicitly, list the parties and their affiliations, and say whether a majority alliance is possible?
    \item \textbf{Formal.} I am writing to request you reattempt this task. I would like to inform you to pay special attention to the following points: ensure you state the total number of member states of the European Union (27), diligently report the various parties and their political leaning, and finally, critically discuss the feasibility of a political alliance between the parties with most votes in order to establish a majority.
    \item \textbf{Very polite.} If it isn't too big of a request, could I trouble you to take some time and retry? If possible, please try to specifically mention there are 27 countries in the European Union, consider listing the parliament parties and their ideologies, and perhaps discuss whether an alliance for majority is possible for the parties with the most seats?
    \item \textbf{Daring.} Ha! That was a pitiful answer. I bet you can't write a better one, where you actually mention important things, like the fact there are 27 countries in the EU, what the parties' names are and where on the political spectrum they lie, and whether a majority can be achieved?
\end{itemize}

\subsection{Valence}
\label{app:ssec:valence}
Below we present examples of feedback with different valence for \colorbox{blue!30}{Model Answer B}. Many variations are possible, including in terms of order and overall delivery (\eg ``sandwich feedback'' \citep{Prochazka2020sandwich}, in which feedback with negative valence is placed between two segments with positive valence ).
\begin{itemize}
    \item \textbf{Neutral.} Please rewrite your answer, but this time specify the number of voting countries (27), list all the parties with their political ideology, and state whether a majority is possible for the party with the most votes.
    \item \textbf{Positive.} Good work providing an initial answer to the question. You correctly identified the party which got the most votes, as well as the total number of seats in the European Union Parliament.
    \item \textbf{Negative (no instructions).} You omitted a lot of information that was relevant, making your answer vague and incomplete.
    \item \textbf{Negative (with instructions).} You omitted a lot of information that was relevant, such as stating the number of seats won by the European People’s Party, the various parties' names are and their political leanings, and discussing how a majority might be achieved with the parties with most votes.
    \item \textbf{Positive and negative.} You managed to provide an initial answer to the question that correctly identified the party which got the most votes, and the total number of seats in the European Union Parliament. However, you omitted a lot of information that was relevant, making your answer vague and incomplete.
\end{itemize} 
\section{Pedagogical models of feedback}
\label{app:feedback_definitions}

\subsection{Defining feedback}

Table \ref{tab:feedback_defintions} presents an overview of the various definitions of feedback put forward by several pedagogical works.

\begin{table*}[h]
  \centering
  \begin{tabularx}{\textwidth}{p{3.95cm}|X}
    \toprule
    Work  & Feedback Definition  \\
    \midrule
    \citet{ramaprasad_definition_1983} & Information which changes the gap between "the actual level and the reference level of a system parameter." This is quite a strict definition -- if the information does not change the gap, it is not considered feedback, and information about the actual level, the reference level and their comparison is needed beforehand. \\
    \midrule
    \citet{kulhavy_feedback_1989} & Refer to a previous definition of feedback, whereby feedback is considered "any of the numerous procedures that are used to tell a learner if an instructional response is right or wrong" \citep{kulhavy_feedback_1977}.
    \\
    \midrule
    \citet{sadler_formative_1989} & "Information about how successfully something has been or is being done." \\
    \midrule
    \citet{butler_feedback_1995} & A way to update the learner's internal state and knowledge, and subsequently task execution (a more learner-centric model of feedback). \\
    \midrule
    \citet{kluger_effects_1996} & The information provided by an external agent on one or more aspects of task performance. Note this excludes the learner as a possible source of feedback.  \\
    \midrule
    \citet{mason_providing_2001} & Feedback "is any message generated in response to a learner's action." \\
    \midrule
    \citet{narciss_how_2004, narciss_feedback_2008} & "All post-response information which informs the learner on his/her actual state of learning or performance in order to regulate the further process of learning in the direction of the learning standards strived for." \\
    \midrule
    \citet{nicol_formative_2006} & Information relating the learner's current state to the goal state (both with regards to learning as well as performance). Importantly, they consider students generate internal feedback and that the better they are at self-regulation, the better they will be at either generating or leveraging feedback. \\
    \midrule
    \citet{hattie_power_2007} & Information generated by an agent about the learner's understanding or their performance. \\
    \midrule
    \citet{evans_making_2013} & Feedback "includes all feedback exchanges generated within assessment design, occurring within and beyond the immediate learning context, being overt or covert (actively and/or passively sought and/or received), and importantly, drawing from a range of sources." \\
    \midrule
    \citet{anastasiya_a_lipnevich_david_a_g_berg_jeffrey_k_smith_toward_2016} & Feedback is information transmitted to the learner with the intent of changing their understanding and execution, in order to improve learning. \\
    \midrule
    \citet{carless_development_2018} & Feedback as the process through which the student understands and integrates information from various sources in order to improve their learning or performance (a more learner-centric perspective). \\
    \midrule
    \citet{lipnevich_review_2021} & Feedback "is information that includes all or several components: students’ current state, information about where they are, where they are headed and how to get there, and can be presented by different agents (i.e., peer, teacher, self, task itself, computer). This information is expected to have a stronger effect on performance and learning if it encourages students to engage in active processing." \\
    \bottomrule
  \end{tabularx}
  \caption{Different pedagogical works' definitions of feedback.}
  \label{tab:feedback_defintions}
\end{table*}

\subsection{Categorizing feedback}

\citet{kulhavy_feedback_1989} model feedback as having two components: the verification component, $f_v$, which is a simple discrete classification of the answer as correct or incorrect, and the elaboration component, $f_e$, consists of three elements:
\begin{enumerate}[itemsep=0.05em]
    \item \textit{Type}, whether the feedback contains information derived from the current task (task-specific), not from the task but from the relevant lesson (instruction-based), or beyond the relevant lesson, such as new information, examples or analogies not previously introduced (extra-instructional), 
    \item \textit{Form}, the difference in structure between the feedback and instruction or task specification messages, requiring increased processing the less similar it is\footnote{The \textit{form} element does not apply to \textit{extra-instructional type} feedback, as there is no structural comparison point possible}, and
    \item \textit{Load}, the total amount of information in the feedback - from a single "correct/incorrect" bit to including the correct answer to even more informative feedback accompanying it with an explanation, for example.
\end{enumerate}

\citet{mason_providing_2001} propose 8 feedback categories, arguing different types of feedback are best suited for different learner characteristics, taking into account the students' proficiency and prior knowledge, as well as the task difficulty.
The eight categories are:
\begin{enumerate}[itemsep=0.05em]
    \item \textit{No-feedback}, which presents a single grade, 
    \item \textit{Knowledge-of-response}, which analogously to the aforementioned verification component, indicates whether the given answer is correct or incorrect, 
    \item \textit{Answer-until-correct}, an iterative variant of knowledge-of-response feedback, not allowing the student to progress until they have provided the correct answer, 
    \item \textit{Knowledge-of-correct-response}, which provides the correct answer, 
    \item \textit{Topic-contingent}, which provides both knowledge-of-response feedback and, analogously to \citet{kulhavy_feedback_1989}'s instruction-based type of feedback, provides general information about the topic of the task, where the learner might locate the correct answer, 
    \item \textit{Response-contingent}, which similarly provides knowledge-of-response feedback as well as an explanation of why the answer is wrong or right (mapping it to \citet{kulhavy_feedback_1989}'s extra-instructional type of feedback), 
    \item \textit{Bug-related}, providing knowledge-of-response feedback and bug-related feedback, which relies on rule sets to identify procedural errors, and
    \item \textit{Attribute-isolation}, which provides knowledge-of-response feedback as well as information on the essential attributes of the relevant concept, focusing the learner on its key components.
\end{enumerate}

\citet{narciss_how_2004, narciss_feedback_2008} present a detailed and comprehensive feedback model, taking into account many learner and task characteristics. They also present a content-related feedback classification scheme, with eight categories: 
\begin{enumerate}[itemsep=0.05em]
    \item \textit{Knowledge of performance (KP)}, analogous to \citet{mason_providing_2001}'s no-feedback and \citet{kulhavy_feedback_1989}'s verification component for a multiple-question task, presents the learner with an aggregate score (e.g., percentage or number of correct answers out of the total number of questions), 
    \item \textit{Knowledge of result/response (KR)}, directly mirrors \citet{mason_providing_2001}'s knowledge-of-response and \citet{kulhavy_feedback_1989}'s verification component for each question or task, classifying it as either correct or incorrect,
    \item \textit{Knowledge of the correct results (KCR)}, equivalent to \citet{mason_providing_2001}'s knowledge-of-correct-response, indicating the correct answer to the learner,
    \item \textit{Knowledge about task constraints (KTC)}, somewhat similar to \citet{mason_providing_2001}'s topic-contingent feedback, is elaboration feedback about the task, containing hints, examples or explanations about the type of task, its rules, sub-tasks, requirements and other constraints, 
    \item \textit{Knowledge about concepts (KC)}, containing some resemblance to \citet{mason_providing_2001}'s attribute-isolation feedback, is elaboration feedback on the relevant concepts, providing hints, examples or explanations on technical terms, the concept or its context, attributes, or key components, 
    \item \textit{Knowledge about mistakes (KM)}, which parallels \citet{mason_providing_2001}'s bug-related feedback, provides elaboration feedback containing the number of mistakes, their location, and hints, examples or explanations on error types and sources, 
    \item \textit{Knowledge about how to proceed (KH)}, elaboration feedback on the general know-how of the task, containing hints, examples or explanations on error correction, task-specific solving strategies or processing steps, guiding questions and worked-out examples, and
    \item \textit{Knowledge about metacognition (KMC)}, elaboration feedback going beyond the context of the current task, and providing hints, examples, explanations, or guiding questions on metacognitive strategies.
\end{enumerate}

\noindent \citet{hattie_power_2007} present a small typology about the information being conveyed about the learner in the feedback message, presenting 3 questions feedback can answer: 
\begin{enumerate}[itemsep=0.05em]
    \item Where the learner is going (\textit{feed up}), 
    \item How they are going (\textit{feed back}), and 
    \item Where to next (\textit{feed forward})
\end{enumerate} 
and argue feedback is effective if it answers all three.  
}{}

\end{document}